\pdfoutput=1
\documentclass[11pt]{article}

\usepackage[final]{acl}

\usepackage{times}
\usepackage{latexsym}
\usepackage{multirow}
\usepackage{booktabs}
\usepackage{amsmath}
\usepackage{amssymb}
\usepackage{longtable}
\usepackage{float}
 \usepackage{hyperref}
\usepackage[T1]{fontenc}
\usepackage[utf8]{inputenc}

\usepackage{microtype}

\usepackage{inconsolata}

\usepackage{graphicx}

\def\ie{\emph{i.e.}}
\def\eg{\emph{e.g.}}

\title{X-LeBench: A Benchmark for Extremely Long Egocentric Video Understanding}

\author{%
  Wenqi Zhou$^1\thanks{Equal contribution.}$ 
  \quad Kai Cao$^{2*}$ 
  \quad Hao Zheng$^1$ 
  \quad Yunze Liu$^6$ 
  \quad Xinyi Zheng$^1$\\
  \quad \textbf{Miao Liu}$^{4,5}$
  \quad \textbf{Per Ola Kristensson}$^3$
  \quad \textbf{Walterio Mayol-Cuevas}$^1$
  \quad \textbf{Fan Zhang}$^1$ \\
  \quad \textbf{Weizhe Lin}$^{3,6}$\thanks{Corresponding author.} 
  \quad \textbf{Junxiao Shen}$^{1,6\dagger}$
  \\\\
  \textsuperscript{1}University of Bristol, \textsuperscript{2}University of Manchester, \textsuperscript{3}University of Cambridge,\\
  \textsuperscript{4}College of AI, Tsinghua University,
  \textsuperscript{5}Meta,
  \textsuperscript{6}Memories.ai Research\\
  {\bf Code}: \href{https://github.com/X-Intelligence-Labs/X-LeBench}{https://github.com/X-Intelligence-Labs/X-LeBench}
}

\begin{document}
\maketitle
\begin{abstract}
Long-form egocentric video understanding provides rich contextual information and unique insights into long-term human behaviors, holding significant potential for applications in embodied intelligence, long-term activity analysis, and personalized assistive technologies. However, existing benchmark datasets primarily focus on single, short (\eg, minutes to tens of minutes)  to moderately long videos, leaving a substantial gap in evaluating extensive, ultra-long egocentric video recordings. 
To address this, we introduce X-LeBench, a novel benchmark dataset meticulously designed to fill this gap by focusing on tasks requiring a comprehensive understanding of extremely long egocentric video recordings. 
Our X-LeBench develops a life-logging simulation pipeline that produces realistic, coherent daily plans aligned with real-world video data. This approach enables the flexible integration of synthetic daily plans with real-world footage from Ego4D—a massive-scale egocentric video dataset covers a wide range of daily life scenarios—resulting in 432 simulated video life logs spanning from 23 minutes to 16.4 hours. 
The evaluations of several baseline systems and multi-modal large language models (MLLMs) reveal their poor performance across the board, highlighting the inherent challenges of long-form egocentric video understanding, such as temporal localization and reasoning, context aggregation, and memory retention, and underscoring the need for more advanced models. 
\end{abstract}

\begin{figure*}[!th]
  \centering
   \includegraphics[width=\linewidth]{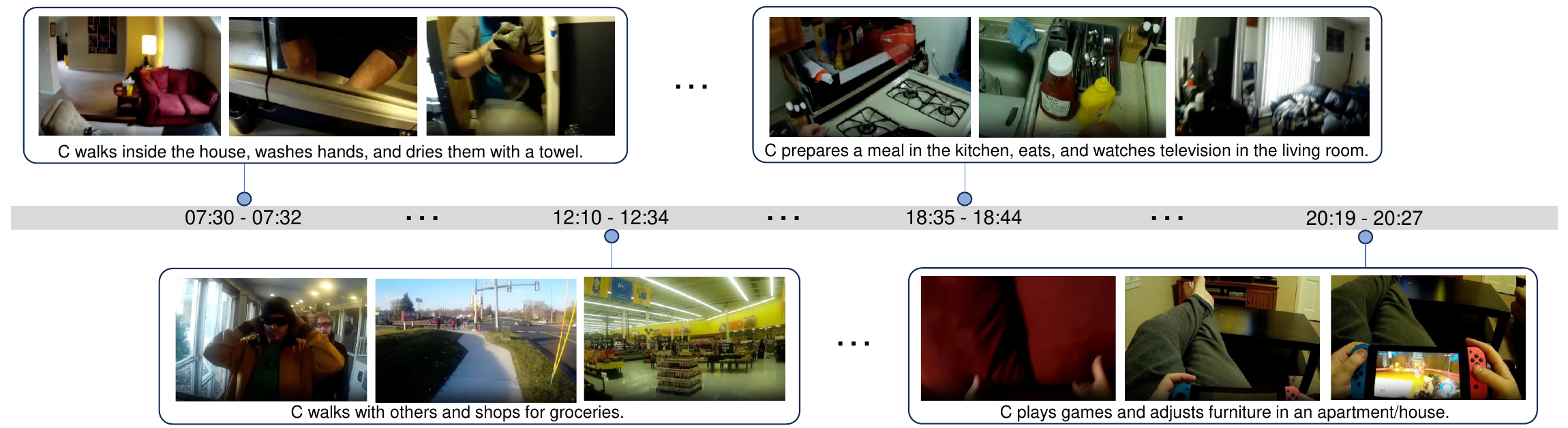}

   \caption{Example of generated video life logs in X-LeBench. Generated video life logs consist of multiple videos with corresponding timestamps. The visualization shows the time organization and content allocation of data.}
   \vspace{-5mm}
   \label{fig:example}
\end{figure*}

\section{Introduction}

Understanding long-form egocentric videos captured from a first-person perspective over extended periods holds significant potential for advancing various domains such as embodied intelligence, long-term activity analysis, and personalized assistive technologies \citep{plizzari2024outlook,lv2024aria,park2016egocentric}. These videos provide rich contextual information and unique insights into human behaviors as they unfold naturally throughout the day. The ability to analyze such extensive recordings is key to developing more personalized agent systems that can construct long-term memory, anticipate user needs, and interact seamlessly in real-world settings~\cite{lin2022egocentric,jia2022egotaskqa,pramanick2023egovlpv2}.

However, most existing datasets~\cite{soomro2012ucf101,caba2015activitynet,kay2017kinetics,li2020hero,xiao2021next,miech2019howto100m}, predominantly feature short individual video clips captured from third-person views, making them insufficient for understanding the continuous, nuanced context of daily human activities or for in-depth human-centered research that requires a first-person perspective.
While more recent benchmarks and datasets focus on egocentric video understanding ~\cite{damen2018scaling,damen2022rescaling,grauman2022ego4d,zhu2023egoobjects,mangalam2023egoschema,grauman2024ego, lv2024aria,fan2025benchmarks}, they remain limited in capturing continuous, long-term human activities. For instance, Ego4D~\cite{grauman2022ego4d} features massive-scale egocentric videos covering a wide range of daily activities, with video durations ranging from 5 seconds to 7 hours, yet subsequent research~\cite{mangalam2023egoschema,grauman2024ego,islam2024video,rodin2024action,chandrasegaran2024hourvideo} often focuses on isolated clips or recordings, falling short of capturing the full scope of long-term daily human activities. 
This limitation hinders the evaluation of models designed to process ultra-long video streams. In particular, it restricts the ability to challenge current long-form video understanding systems and models that construct long-term memory from video recordings and retrieve relevant information in response to user queries. Without extensive, continuous contextual data, evaluating the robustness of their performance becomes difficult.

Creating benchmark datasets that span several hours of egocentric video is necessary but presents significant challenges. Data acquisition is a primary hurdle, requiring participants to wear recording devices for extended periods is labor-intensive and raises privacy concerns. Device limitations, including storage constraints and reliability issues, further complicate continuous video capture. Additionally, annotating long-form videos is time-consuming and prone to annotator fatigue, affecting label accuracy and consistency.

To address these challenges, we introduce X-LeBench, a versatile and scalable benchmark dataset designed for evaluating tasks on extremely long egocentric videos. X-LeBench features a life-logging simulation pipeline that simulates extended video logs by integrating short (seconds) or moderately long (hours) video clips with dynamically generated daily plans. Leveraging large language models (LLMs), it generates realistic, contextually rich schedules aligned with real-world activities based on adjustable input settings. Specifically, this simulation integrates synthetic daily plans with actual footage from Ego4D, then iteratively optimizes the simulation process based on retrieved information, producing video life logs that mirror daily activities in rich contexts with duration extended to dozens of hours. 
A generated sample is shown in Fig.~\ref{fig:example}, showcasing video segments with different timestamps. 
Notably, X-LeBench offers a customizable and scalable design, enabling the synthesis of datasets with various durations and content to accommodate diverse research needs.

Our initial evaluations of baseline systems and multi-modal LLM (MLLM) reveal consistently poor performance on X-LeBench, highlighting the inherent difficulties of long-form egocentric video understanding and underscoring the need for more advanced models capable of interpreting and analyzing ultra-long egocentric videos. 

Our contributions are: (1) We present X-LeBench, the first benchmark dataset that encompasses ultra-long egocentric video recordings; (2) We introduce a novel and customizable pipeline that simulates realistic, hours-long egocentric video life logs by integrating synthetic daily plans with real-world footage; (3) We conduct extensive evaluations of existing models on X-LeBench, exposing significant performance gaps and key challenges for future research.

\section{Related Works }
\label{sec:related}

\subsection{Egocentric Video Benchmarks}
Egocentric video understanding has received increasing attention~\cite{cheng2024egothink,huang2024vinci}. Ego4D is a landmark dataset, offering 3,670+ hours of egocentric footage across 74 locations, facilitating significant progress in the field. 

Related works~\cite{bain2022whisperx,kevin2022egovlp,pramanick2023egovlpv2,islam2024video} extend Ego4D, further exploring various applications of egocentric video understanding. 
EgoSchema~\cite{mangalam2023egoschema} offers an egocentric video question-answering benchmark with over 5,000 curated multiple-choice pairs but focuses on 3-minute clips. The AEA dataset~\cite{lv2024aria} offers multi-modal egocentric data, HourVideo~\cite{chandrasegaran2024hourvideo} curates hour-long videos from Ego4D for evaluating video-language understanding. 
EgoPlan-Bench2~\cite{qiu2024egoplan} assesses planning capabilities of MLLMs. However, these works either lack coherent, continuous, long-term daily-life recordings or face data scale and diversity limitations.

Recently, EgoLife's week-long recordings~\cite{yang2025egolife} offer valuable long-context data, but high costs, operational issues, limited 6 subjects, and an indoor focus restrict its scalability and diversity. In contrast, we developed a novel, cost-effective pipeline to synthesize ultra-long, coherent egocentric video life logs from existing datasets, offering a scalable and extensible alternative and significantly broadening the scope and applicability of long-term egocentric video understanding.

\begin{table}[!h]
\renewcommand{\arraystretch}{0.8}
\centering
\vspace{-1mm}
\caption{The comparison of various benchmarks.}
   \vspace{-2mm}
\scalebox{0.52}{
\begin{tabular}{cccccc}
\toprule
&  &  &   \\[-10pt]
\textbf{Dataset}    & \textbf{\begin{tabular}[c]{@{}c@{}}Avg. Duration\\(mins)\end{tabular}} & \textbf{\begin{tabular}[c]{@{}c@{}}Annotation\\Scheme\end{tabular}} & \textbf{Egocentric}  & \textbf{\#QAs}  & \textbf{\#Data} \\
\hline
&  &   & \\[-6pt]
MVBench & 0.27  &  Auto   &  No  &  4,000  &  3,641 \\
\hline% \cline{2-6}
&  &   & \\[-6pt]
ActivityNet-QA & 1.85    &  Manual  &  No  &  8,000  &  800   \\
\hline
&  &   & \\[-6pt]
EgoSchema & 3     &  Auto\&Manual &   Yes & 5,063 & 5,063    \\
\hline
&  &   & \\[-6pt]
EgoPlan-Bench2  & up to 5     &  Auto\&Manual &   Yes   & 1,321 & 1,113 \\
\hline% \cline{2-6}
&  &   & \\[-6pt]
MovieChat-1K &  9.4   &  Manual  & No &  1,950  &  130   \\
\hline
&  &   & \\[-6pt]
MLVU &  12   &  Auto\&Manual   & Partial &  2,593  &  757  \\
\hline
&  & & \\[-6pt]
Video-MME (Short) & 1.37   & \multirow{3}{*}{Manual}   & \multirow{3}{*}{No} & \multirow{3}{*}{2,700} & \multirow{3}{*}{900} \\
% \hline
&  & & \\[-6pt]
Video-MME (Medium) & 9.38   &    &      \\
% \hline
&  & & \\[-6pt]
Video-MME (Long) & 39.76  &    &     \\

\hline
&  &   & \\[-6pt]
HourVideo & 45.7  &  Auto\&Manual &   Yes  & 12,976 & 500 \\
\hline
&  &   & \\[-6pt]
InfiniBench & 76.34  &  Auto\&Manual &   No  &  108,200 &  1219 \\
\hline
&  &  & \\[-6pt]
LVBench &  68.35   & Manual &   No &  1,594  &  103 \\
\hline
&  &  & \\[-6pt]
EgoLife &  2,658   & Auto\&Manual &  Yes & 3,000 & 6 \\
\hline
\hline
&  &   & \\[-6pt]
Ours (Short) &  142   &  \multirow{3}{*}{Auto\&Manual}   & \multirow{3}{*}{Yes}  &  \multirow{3}{*}{26,932} & \multirow{3}{*}{432} \\
% \hline
&  &   & \\[-6pt]
Ours (Medium) &  319   &     &    \\
% \hline
&  &   & \\[-6pt]
Ours (Long) &  516   &    &     \\
\bottomrule
\end{tabular}
}
\label{table:comparison}
   \vspace{-5mm}
\end{table}

\subsection{Long-form Video Benchmarks}
The definition of ``long'' in video understanding varies across benchmarks, with durations ranging from minutes to hours.
As shown in Tab.~\ref{table:comparison}, EgoSchema defines 3-minute videos as long using their proposed certificate length, Video-MME~\cite{fu2024video} classifies videos with a length of 30-60 minutes as long. HourVideo and LVBench~\cite{wang2024lvbench} define long videos as 20+ and 30+ minutes, respectively. 
MLVU~\cite{zhou2024mlvu} offers videos of diversified lengths with a 12-minute average length, providing comprehensive evaluation tasks for MLLMs' long video understanding capabilities. InfiniBench~\cite{ataallah2024infinibench} pushes the boundary with 50-minute videos and over 108,000 question-answer pairs, posing significant challenges for leading AI models. 

Despite recent advances, existing datasets remain insufficiently long for evaluating ultra-long video processing and rarely focus on egocentric content. X-LeBench fills this gap by redefining ``long video'' to include multi-hour, contextually consistent egocentric recordings with rich annotations to advance model and system development.

\subsection{LLM-assisted Annotation Scheme} 
Traditional annotation processes require extensive human effort, but advances in LLMs and MLLMs~\cite{achiam2023gpt,yao2024minicpmvgpt4vlevelmllm,Qwen-VL,team2023gemini,touvron2023llama,lin2023video} have facilitated automated annotation in benchmarks. As shown in Tab.~\ref{table:comparison}, several video understanding benchmarks leverage LLMs/MLLMs to streamline annotation process~\cite{mangalam2023egoschema,rawal2024cinepile,patraucean2023perception, li2024mvbench}. For instance, EgoSchema generates question-answer pairs by querying an LLM with human-annotated narrations and tailored prompts.

Given the multi-hour duration of our dataset, manual annotation is impractical and prone to fatigue-induced errors. To address this, we also adopt an automatic and manual annotation scheme. First, we adapt Ego4D's manual annotations to align with our task requirements. 
Next, we use LLMs to consolidate summaries of 5-minute video clips from Ego4D at various granularities, creating single-video, multi-video, and holistic-level summaries (detailed in Appendix~\ref{appen:tasks}).

\section{X-LeBench}
\label{sec:X-LeBench}
In real-world scenarios, continuous egocentric life-logging is often constrained by hardware limitations and privacy concerns, making uninterrupted long-term recordings infeasible.
Consequently, extended life logs must be reconstructed from multiple video segments captured at different times. Existing benchmark datasets lack the structural continuity and task setup required for studying temporally extended, cross-hour behaviors and reasoning.

Building on this insight, we develop the life-logging simulation pipeline, enabling scalable and customizable simulation of hours-long video life logs, while maintaining reality and contextual coherence. We also define tasks that better reflect real-world long-term requirements and go beyond Ego4D’s original scope. The following sections detail our methodology for X-LeBench creation.

\begin{figure*}[ht!]
  \centering
   \includegraphics[width=\linewidth]{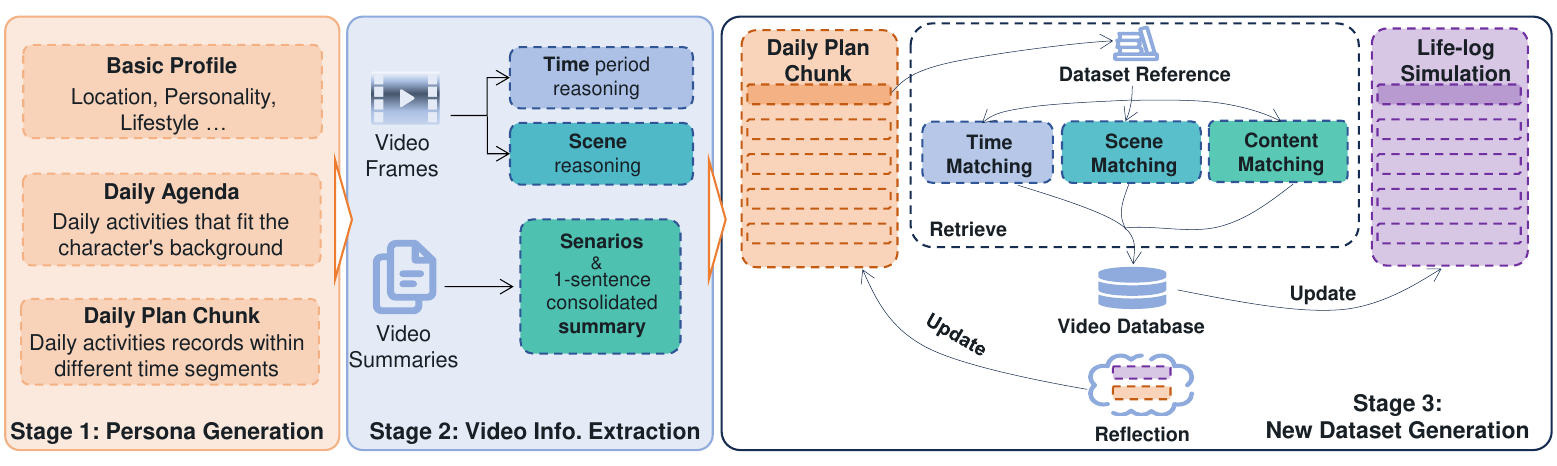}

   \caption{Overview of life-logging simulation pipeline. Stage 1: Generation of personalized persona profiles and daily plan chunks based on predefined parameters. Stage 2: Core information (time, scene, content) extraction of videos in the video selection library. Stage 3: Matching and retrieval of daily plan chunks with videos, and life-log simulations are iteratively refined through reflection, resulting in the final optimized output.}
   \label{fig:simulation}
   \vspace{-5mm}
\end{figure*}

\subsection{Life-logging Simulation Pipeline} 
\label{subsec:sim_pipeline}

As shown in Fig.~\ref{fig:simulation}, we constructed the pipeline by implementing the following three stages:

\noindent\textbf{Stage 1 - Persona Generation.}
To ensure the simulations better reflects the multifaceted nature of real-world human activity, enriching the realism and variability of the resulting dataset, 
we dynamically generate personalized character profiles based on different predefined locations and the Myers-Briggs Type Indicators (MBTI)~\cite{myers1962myers}, generating basic background information including the character's personality traits, lifestyle, hobbies and general daily routines. By incorporating varied character settings, we capture a wide range of behavioral variations that enrich our dataset. This pragmatic choice enables the generation of diverse daily plans and activity patterns, promoting variability across simulations.

Each daily plan is then segmented into time-specific activity chunks, ensuring structured activity distribution.
During this stage, we use GPT-4o~\cite{hurst2024gpt} for character generation, setting 9 different locations based on Ego4D’s demographic data, and 16 MBTI types per location, resulting in 144 diverse persona profiles (profile example is shown in Appendix~\ref{appen:profile}).

\noindent\textbf{Stage 2 - Video Information Extraction.} 
We construct our video selection library by carefully selecting 7852 videos labeled with scenario information and dense clip-level summaries from the Ego4D, excluding redacted content. Then we extract three key attributes from each video, including the time, scene, and the main content. 
Specifically, we use Gemini-1.5-Pro~\cite{team2024gemini} to analyze videos and transform information. To facilitate simulation with reasonable environment settings and lighting conditions, videos are categorized into different scene environments: indoor, outdoor, and mixed, as well as different time periods: daytime, nighttime, twilight, and uncertain (\eg, indoor videos lacking external cues are labeled ``uncertain''). 
Content information is derived from Ego4D scenario tags, and 1-sentence summaries consolidated from 5-minute clip-level summaries from each video. This structured metadata ensures precise alignment between video content and the generated daily plans in the next stage.

\begin{figure*}[ht!]
  \centering
   \includegraphics[width=\linewidth]{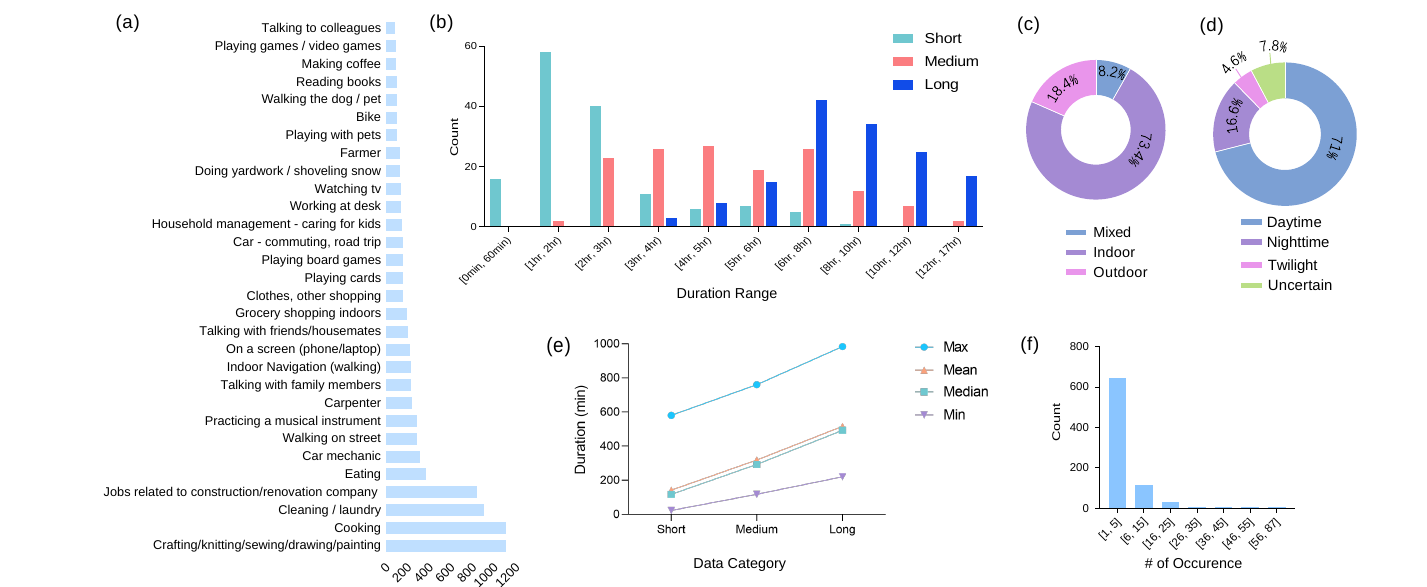}

   \caption{Dataset statistics. (a) Video selection library consists of 7852 videos from Ego4D dataset, covering 135 scenarios, here shows the top 30. (b) X-LeBench has video life-log durations spanning from 23 minutes to 16.4 hours, here shows distribution of duration range across different data categories. (c) and (d) show the distribution of scene and time information categories of videos in video selection library, respectively. (e) Statistical information on duration lengths of different data categories, including minimum, mean, median and maximum values. (f) The distribution of number of videos with different occurrences.}
   \label{fig:statistics}
   \vspace{-5mm}
\end{figure*}

\noindent\textbf{Stage 3 - New Dataset Generation.} 
In this stage, we match the generated daily plan chunks (Stage 1) with videos (Stage 2) to construct coherent video life-log simulations. For each plan chunk, we retrieve a video that aligns with its time, scene, and content, the selection process follows rule below:
\begin{equation}
  \scalebox{0.8}
  {$V ={\underset
  {V\in \mathbf{M_v}}{{\arg\max}}\ S(\mathbf{M_v}  (C_{t}, C_{scene}, C_{scenario}), C_{desc})}$}
  \label{eq:simulation}
\end{equation}
where $V$ is the selected video for chunk $C$, determined by the alignment with $C$'s core information, which is time ($C_t$), scene ($C_{scene}$), scenario ($C_{scenario}$), and 1-sentence summary description ($C_{desc}$). 
$\mathbf{M_v(\cdot)}$ denotes the operation of matching videos based on the provided chunk information. $S(\cdot)$ represents the sentence similarity computation between $C_{desc}$ and 1-sentence summary set of matching videos. 
Specifically, $C_t$ and $C_{scene}$ are inferred from chunk timestamps and content, while $C_{scenario}$ is inferred using the dataset’s predefined scenario list (\ie, the dataset reference in Fig.~\ref{fig:simulation}). 

To efficiently identify the most suitable videos from the large-scale video selection library, selection process follows a coarse-to-fine retrieval strategy. We first apply a coarse-grained filtering step, \ie, the operation $\mathbf{M_v}  (C_{t}, C_{scene}, C_{scenario})$, a rule-based matching to narrow down the candidate video set. Finally, to refine the selection, we calculate the sentence similarity between $C_{desc}$ and the summaries of the candidate video set $\mathbf{M_v}$, ensuring alignment across time, scene, and content dimensions.

In addition, to enhance the contextual reasonableness and coherence of overall video life logs, we preferentially select videos recorded in the same location for each simulation. Furthermore, an iterative optimizing strategy is adopted, \ie, after each matching of a daily plan chunk, we update the life-log simulation memory with the information of selected video, require LLM uses this context to reasonably adjust and update the subsequent daily plan chunks, ensuring that the overall simulation maintains logical coherence and contextual alignment. Details can be found in Appendix~\ref{appen:satge3}.

\noindent\textbf{Statistics.}
Our life-logging simulation pipeline allows for flexible customization by inputting various locations and MBTI types, generating diverse persona profiles and simulations. We can customize different numbers of chunks to get video life logs of various video compositions. 
In this work, we set three different chunk numbers for simulation, namely 4, 9, and 15 (corresponding to short, medium, and long life-log categories) for each location-MBTI combination generated persona, thus obtaining 432 extremely long video life logs. 
To assess dataset quality, we compare our method with a randomly sampled baseline (Appendix~\ref{appen:additional_results}), our method retrieves videos with contextualized timestamps and more consistent content that closely resemble real-life recorded scenes. We also conduct a five-point human evaluation on realism and contextual consistency. As summarized in Tab.~\ref{table:dataset}, Fig.~\ref{fig:statistics} (b) and (e), our dataset achieved scores over 4 across all categories, demonstrating strong alignment with real-world activity patterns and logical continuity. Evaluation details are provided in the Appendix~\ref{appen:quality}. Also, the dataset exhibits diverse duration distributions: Short data are mostly 1-2 hours; Medium data are mostly 4-5 hours; And long data are mostly 6-8 hours. With the average duration of 2.37, 5.32 and 8.6 hours. 

\begin{table}[!h]
\renewcommand{\arraystretch}{0.8}
\centering
\caption{Statistics of dataset duration length and quality.}
\vspace{-2mm}
\scalebox{0.65}{
\begin{tabular}{cccccc}
\toprule
% &  &  &  & &  \\[-10pt]
\textbf{Life-log Category}    & \textbf{\begin{tabular}[c]{@{}c@{}}Max\\(min)\end{tabular}} & \textbf{\begin{tabular}[c]{@{}c@{}}Mean\\(min)\end{tabular}} & \textbf{\begin{tabular}[c]{@{}c@{}}Min\\(min)\end{tabular}} & \textbf{Realism} & \textbf{\begin{tabular}[c]{@{}c@{}}Contextual\\Consistency\end{tabular}}\\
\midrule
% &  &  &   \\[-6pt]
Short & 580               & 142        & 23    & 4.71 & 4.50   \\
% \cline{2-6}
&  &  &  \\[-6pt]
Medium & 760  &     319 &       118  & 4.26 & 4.37      \\
% \cline{2-6}
&  &  &   \\[-6pt]
Long & 984      & 516       & 220   & 4.41 & 4.02   \\                     
\bottomrule
\end{tabular}
}
\label{table:dataset}
\vspace{-3mm}
\end{table}

Our video selection library contains 7852 videos, covering a wide range of daily life scenarios and scenes across different times of the day, a total of 135 scenarios are covered. Fig.~\ref{fig:statistics} (a), (c) and (d) illustrate the top 30 most frequent scenarios, time distributions, and scene types. In addition, due to the commonality of human activities (\eg, eating, cooking), certain videos are selected multiple times. Fig.~\ref{fig:statistics} (f) shows that most videos appear fewer than five times, with a small number of videos appearing more than 5 times, reflecting the diversity of our simulations while preserving the representativeness of repeated contexts.

\begin{table}[!h]
\vspace{-2mm}
\renewcommand{\arraystretch}{1}
\centering
\caption{The number of annotations.}
   \vspace{-3mm}
\scalebox{0.55}{
\begin{tabular}{cccc}
\toprule
\multicolumn{1}{c|}{}&  \multicolumn{1}{c|}{}&  \multicolumn{1}{c|}{}& \multicolumn{1}{c}{} \\[-10pt]
\multicolumn{1}{c|}{\textbf{Object-related Retrieval}}    & \multicolumn{1}{c|}{\textbf{\begin{tabular}[c]{@{}c@{}}People-related\\ Retrieval\end{tabular}}} & \multicolumn{1}{c|}{\textbf{Action Counting}} & \multicolumn{1}{c}{\textbf{Summary Ordering}}  \\ \cline{1-4}
\multicolumn{1}{c|}{}&  \multicolumn{1}{c|}{}&  \multicolumn{1}{c|}{}& \multicolumn{1}{c}{} \\[-10pt]
\multicolumn{1}{c|}{1444}                                       & \multicolumn{1}{c|}{583}                                         & \multicolumn{1}{c|}{5295}                             & \multicolumn{1}{c}{4032}   \\
\bottomrule

&  &  &   \\[-6pt]
\toprule
\multicolumn{1}{c|}{}&  \multicolumn{3}{c}{} \\[-10pt]
\multicolumn{1}{c|}{\multirow{2}{*}{\textbf{Moment Retrieval}}} & \multicolumn{3}{c}{\textbf{Summarization}} \\
\cline{2-4}
\multicolumn{1}{c|}{}& \multicolumn{1}{c|}{} & \multicolumn{1}{c|}{} & \multicolumn{1}{c}{} \\[-10pt]
 \multicolumn{1}{c|}{} & \multicolumn{1}{c|}{single-video}    & \multicolumn{1}{c|}{multi-video} & \multicolumn{1}{c}{holistic}\\
\hline
\multicolumn{1}{c|}{}                     & \multicolumn{1}{c|}{}                & \multicolumn{1}{c|}{}    & \multicolumn{1}{c}{}\\[-10pt]
\multicolumn{1}{c|}{9869}                 & \multicolumn{1}{c|}{4032}            & \multicolumn{1}{c|}{1245}        & \multicolumn{1}{c}{432}  \\
\bottomrule
\end{tabular}
}
\label{table:tasks}
\vspace{-5mm}
\end{table}

\begin{table*}
% \ra{1.3}
\centering
\resizebox{\textwidth}{!}{
\begin{tabular}{@{}lll@{}}\toprule
\textbf{Temporal Localization} &  & \\ \midrule
Object-related Retrieval & Q: What did I put on the table in the record provided for the 21:56 - 22:25 time period?  & A: [22:01:55 - 22:02:03], [22:02:20 - 22:02:42], ...\\
People-related Retrieval & Q: Who did I talk to in the living room in the record provided for the 21:56 - 22:25 time period? & A: [22:02:35 - 22:02:39], [22:03:30 - 22:03:50], ...\\
Moment Retrieval & Q: When did I $\langle$use phone$\rangle$ in the record provided for the 21:56 - 22:25 time period? &A: [22:15:00 - 22:15:06], [22:18:49 - 22:19:00], ...\\
\midrule
\textbf{Summarization} & & \\ \midrule
Single-video Summarization & Q: Summarize the activities performed in the recordings provided for the time period 22:13 - 22:25. & A: C interacts with others and writes in a house and a studying room.\\
Multi-video Summarization & Q: Summarize the activities performed in the recordings provided in the morning (before 12:00). & A: C engages in various activities, including preparing meals, riding...\\
Holistic Summarization & Q: Summarize the activities performed in all provided life-log recordings. & A: C engages in a variety of activities throughout the day, including\\
& \hspace{1em}  & \hspace{1em} preparing meals, riding, visiting restaurants, ...\\
\midrule
\textbf{Counting} &\\ \midrule
Action Counting & Q: In the record provided for the 07:25 - 07:48 time period, how many times have each of the & A: wipe soap: 2; remove cheese: 1; ... \\
& \hspace{1em} actions in the following list been performed in the 0-28 second record of the period?
\\
& \hspace{1em} Action list: 1. wipe soap, 2. remove cheese, ... \\
\midrule
\textbf{Ordering} &\\ \midrule
Summary ordering & Q: Please rank the following summaries of camera wearer C's activities in order of presentation of & A: Correct order of the summaries: \\
& \hspace{1em} the life-log recordings. & \hspace{1em}  7, 2, 11, 5, 8, 1, 6, 3, 4, 0, 12, 9, 13, 10, 14.\\
& \hspace{1em} Summary 0: C climbs, interacts with others, and observes climbing activities in various indoor...\\
& \hspace{1em} ...\\
& \hspace{1em} Summary 14: C interacts with others and writes in a house and a studying room.\\
\bottomrule
\end{tabular}
}
\caption{Tasks in this benchmark and their corresponding examples. \textbf{Q} and \textbf{A} denote query and answer examples.}
\vspace{-5mm}
\label{table:task_examples}
\end{table*}

\subsection{Benchmark Tasks}
\label{subsection:tasks}
As shown in Tab.~\ref{table:task_examples}, 
X-LeBench introduces a suite of evaluation tasks specifically designed for daily-life long-form egocentric videos. These tasks encompass object-, people-, and moment-related temporal localization, multi-level summarization, action counting, and summary ordering. Details and examples are in Appendix~\ref{appen:tasks}.

The full-length video is presented to the system only once before querying. The system is required to extract features or frames from the extremely long video and store them in a buffer, responding to query based solely on the stored information. This approach presents a significant challenge to current video understanding systems, which are typically designed to process short videos and thus cannot store information in the presented videos in buffers for future use. 
Furthermore, as video duration increases, answers to queries may recur across multiple time segments. To address this, we explicitly annotate the queried period for each task, ensuring precise and unambiguous retrieval, while mitigating annotation gaps inherent in single-video datasets. After generating the life-log footage, we adapt corresponding Ego4D annotations for our novel task designs. Tab.~\ref{table:tasks} summarizes the number of annotations per task.

\begin{table*}[!ht]
\centering
\renewcommand{\arraystretch}{1.3} % Adjust row spacing
\resizebox{1\textwidth}{!}{
\begin{tabular}{l|c|c|c|c|c|c|c|c|c|c|c}
\toprule
\multirow{2}{*}{\textbf{Method}} & \multirow{2}{*}{\textbf{\begin{tabular}[c]{@{}c@{}}Data\\Category\end{tabular}}} & \multirow{2}{*}{\textbf{\begin{tabular}[c]{@{}c@{}}Object-related\\Retrieval (\%)\end{tabular}}} & \multirow{2}{*}{\textbf{\begin{tabular}[c]{@{}c@{}}People-related\\Retrieval (\%)\end{tabular}}} & \multirow{2}{*}{\textbf{\begin{tabular}[c]{@{}c@{}}Moment\\Retrieval (\%)\end{tabular}}} & \multicolumn{3}{c|}{\textbf{Summarization (10)}} & \multirow{2}{*}{\textbf{\begin{tabular}[c]{@{}c@{}}AVG. Temporal\\Localization (\%)\end{tabular}}} & \multirow{2}{*}{\textbf{\begin{tabular}[c]{@{}c@{}}AVG.\\Summarization (10)\end{tabular}}} & \multirow{2}{*}{\textbf{\begin{tabular}[c]{@{}c@{}}Action\\Counting (\%)\end{tabular}}} & \multirow{2}{*}{\textbf{\begin{tabular}[c]{@{}c@{}}Ordering\\(\%)\end{tabular}}} \\ \cline{6-8}
                &                       &                     &                    &                    & \textbf{single}     & \textbf{multi-video} & \textbf{holistic}     &                    &                    &                    &                  \\ \hline
\multirow{3}{*}{Gemini-1.5-Flash} & Short & 1.67 & 0.00 & 10.26 & 5.12 &
3.02 &
4.90 & 4.20 & 4.36 & 13.87 & 93.70 \\ \cline{2-12}
 & Medium & 1.92 & 6.90 & 10.53 & 4.82 &
3.12 &
5.24 & 5.88 & 4.46 & 16.74 & 44.66 \\ \cline{2-12}
 & Long & 1.75 & 11.11 & 8.33 & 4.57 &
3.44 &
5.33 & 6.45 & 4.43  & 16.76 & 24.40 
\\
\hline
\hline
\multirow{3}{*}{Socratic Models} & Short & 8.33 & 25.00 & 17.95 & 6.80 &
3.69 &
6.47 & 14.28 & 5.67 & 9.53 & 85.59 \\ \cline{2-12}
 & Medium & 5.77 & 20.69 & 18.42 & 6.64 &
4.09 &
6.41 & 13.45 & 6.04 & 14.04 & 55.63 \\ \cline{2-12}
 & Long & 12.28 & 22.22 & 10.83 & 5.82 &
4.37 &
6.15 & 11.83 & 5.61 & 19.85 & 23.33 
\\
\hline
\hline
\multirow{3}{*}{Retrieve - Socratic} & Short & 8.33 & 25.00 & 17.95 &6.88 &
3.79 &
6.53 & 14.28 & 5.75 & 9.92 & 87.67 \\ \cline{2-12}
 & Medium & 9.62 & 17.24 & 7.89 & 7.01 &
4.30 &
6.56 & 10.92 & 6.36 & 9.38 & 54.55 \\ \cline{2-12}
 & Long & 10.53 & 66.67 & 14.17 &6.84 &
4.69 &
6.48 & 15.59 & 6.48 & 9.77 & 24.17 
\\ \hline
\hline
Gemini-1.5-Flash & All & 1.77 & 5.17 & 9.14 & 4.73 &
3.20 &
5.16 & 5.66 & 4.43 & 16.44 & 41.96
\\
\hline
Socratic Models & All & 8.87 & 22.41 & 13.70 & 6.22 &
4.06 &
6.34 & 12.97 & 5.76 & \textbf{17.15} & 42.61
\\
\hline
Retrieve - Socratic & All & \textbf{9.47} & \textbf{27.59} & \textbf{13.71} & \textbf{6.90} &
\textbf{4.28} &
\textbf{6.52} & \textbf{13.92} & \textbf{6.30} & 9.67 & \textbf{43.01}
\\ 
\bottomrule
\end{tabular}}
\caption{Evaluation results on X-LeBench. Including temporal localization tasks (object-related, people-related and moment retrieval), summarization tasks (single-video, multi-video and holistic level), action counting and summary ordering task. AVG. temporal localization and summarization is the average performance of all temporal localization tasks and summarization tasks, respectively.}
\label{tab:performance}
\end{table*}

\begin{figure*}[ht]
  \centering
   \includegraphics[width=\linewidth]{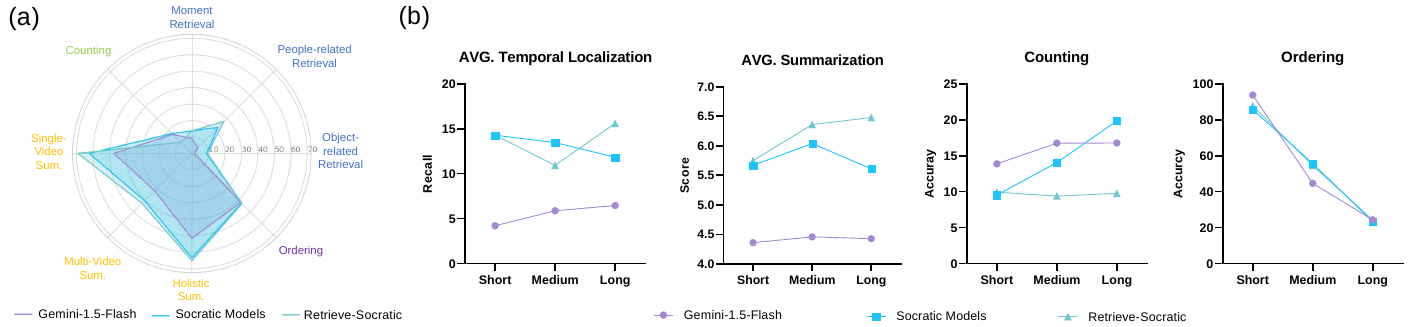}

   \caption{Performance comparison on X-LeBench. (a): Radar chart showing the overall performance of each method across all tasks. For fair comparison across tasks, scores for summarization are scaled by a factor of 10 (maximum 100). (b): The performance comparison across different data categories for different tasks.}
   \label{fig:results}
   \vspace{-5mm}
\end{figure*}

\section{Experiments}
\label{sec:experiments}

\subsection{Settings}
X-LeBench includes various task types, as outlined in Sec.~\ref{subsection:tasks}, encompassing four categories: temporal localization, summarization, counting, and ordering. These tasks are further divided into eight subtasks, with their respective input-output formats detailed in Tab.~\ref{table:task_examples}.
Given the interleaved, ultra-long chronological context (Fig.~\ref{fig:example}), we uniformly construct time-stamped prompts as the long-context input following the format shown in Fig.~\ref{fig:context}.

To reflect real-world usability, tasks require free-form textual outputs instead of closed-set multiple-choice answers. This approach increases practical usability while significantly raising the complexity of tasks, as it demands more nuanced understanding and reasoning. To ensure the evaluability of outputs, the prompts provide extended context along with task-specific instructions and output format requirements for each task type, standardizing the expected responses. 
For temporal localization and summarization, each query is independently evaluated, while for counting and ordering, we aggregate the query contents of each data entry to action or order lists for a unified assessment. Notably, due to computational cost and time associated with analyzing ultra-long contexts, along with their overall poor performance and limited analytical value—we test only a limited set on temporal localization. 

Overall, X-LeBench presents significant challenges for current video understanding systems, which are often limited by short context handling, weak temporal reasoning, and poor long-horizon memory retrieval. To evaluate performance under these constraints, we test three representative multi-modal approaches to handle ultra-long contexts:

(1) Gemini-1.5-Flash~\cite{team2024gemini}: A native multi-modal model excels in handling ultra-long contexts and is trained jointly on multi-modal data. It is used in an end-to-end manner by uniformly sampling 1200 frames per data, with a temperature of 0.1.

(2) Socratic Models~\cite{zeng2022socraticmodels}: Most advanced multi-modal systems cannot process ultra-long videos. Inspired by related works~\cite{chandrasegaran2024hourvideo,zeng2022socraticmodels}, we divide videos into 30-second segments and generate captions for these segments using the Qwen-VL-7B model~\cite{Qwen2-VL}, captioning segments at 1 frame per second and resolution settings of $560\times420$. The captions are then aggregated with timestamps as life-log records, forming the input for executing long-context understanding tasks. The Gemini-1.5-Flash model is used for performing tasks on these textual life log records, with the same parameters as in method (1).

(3) Retrieve-Socratic: Considering the inherent temporal nature of our dataset, we propose an enhanced version of the Socratic Models approach. This method incorporates a rule-based filtering mechanism to extract only task-relevant time segments, reducing irrelevant temporal data. By narrowing the contextual focus, this approach aims to improve efficiency and relevance in long-context understanding. The extracted contextual information is then processed using the same Socratic Models' pipeline and parameter settings as in method (2). 

We also evaluate the LongVU~\cite{shen2024longvu}, an open-source method for long-form video understanding. Detailed results are in Appendix~\ref{appen:additional_results}.

\subsection{Results}
As shown in Fig.~\ref{fig:results} and Tab.~\ref{tab:performance}, we evaluate the performance of three methods across short, medium, long, and all data categories. The ``all'' category represents an aggregated evaluation of the former 3 data types, offering a more comprehensive and holistic assessment. The evaluation metrics for each task are detailed in Appendix~\ref{appen:tasks}. 

\noindent\textbf{Overall Performance.} 
As shown in last three rows of Tab.~\ref{tab:performance} and Fig.~\ref{fig:results} (a), the Retrieve-Socratic method demonstrates superior performance in most tasks. Due to input token limitations, longer videos require larger frame sampling intervals, resulting in increased information loss. Consequently, both Socratic and Retrieve-Socratic methods significantly outperform Gemini-1.5-Flash on temporal localization and summarization tasks. Specifically, Retrieve-Socratic achieves an average improvement of 8.26\% in recall for temporal localization and 1.87 points higher scores in summarization. The underperformance of Gemini-1.5-Flash in temporal localization reflects its difficulty in fine-grained temporal reasoning under long-context constraints. Additionally, Retrieve-Socratic further improves upon the basic Socratic approach by narrowing the temporal context, leading to notable gains in both temporal localization and summarization tasks. In holistic summarization and ordering, where the queried information spans the entire video, the performance of Socratic and Retrieve-Socratic methods remains comparable, demonstrating their robustness in global reasoning. 

Interestingly, our analysis of the specific results indicates that the Retrieve-Socratic method tends to output 0 more frequently in the counting task. This leads to noticeably poorer performance compared to other methods, suggesting that excessive context reduction may introduce unintended biases in certain tasks.

\noindent\textbf{Impact of Data Types.} 
As illustrated in Fig.~\ref{fig:results} (b) and Tab.~\ref{tab:performance}, the performance trends across data types are not uniform. These trends may be influenced by varying task-specific objectives and understanding burden imposed by different data types on each method. 
Overall, both Socratic and Retrieve-Socratic demonstrate more stable performance trends across data types compared to Gemini-1.5-Flash. This suggests that the processing capability of textual information is more robust when dealing with such long-context information. 
For the ordering task, all methods achieve strong performance on short videos (accuracy exceeding 85\%). However, as video length increases, performance across all methods declines significantly, with accuracy dropping below 25\% for long data. 

\noindent\textbf{Impact of Temporal Structure.} 
To explore how the complexity of temporal structures impacts the difficulty of core tasks involving temporal understanding, we conducted experiments on a ``shorter'' split of our dataset, where each video life log consists of only two videos. This setup enables controlled analysis of temporal reasoning under reduced structural complexity. The comparative results with the ``short'' split are presented in Tab.~\ref{tab:shorter_split}.

\begin{table}[H]
\centering
\scalebox{0.568}{
\begin{tabular}{lcccc}
\toprule
\textbf{Method} & \textbf{\begin{tabular}[c]{@{}c@{}}Single-video\\Sum. (10)\end{tabular}} & \textbf{\begin{tabular}[c]{@{}c@{}}Multi-video\\Sum. (10)\end{tabular}} & \textbf{\begin{tabular}[c]{@{}c@{}}Ordering\\(\%)\end{tabular}} & \textbf{\begin{tabular}[c]{@{}c@{}}AVG.\\Temporal Loc. (\%)\end{tabular}} \\
\midrule
Gemini-1.5-Flash & \begin{tabular}[c]{@{}c@{}}5.52 \\ [-0.3em] ($\uparrow$ 0.40)\end{tabular} 
       & \begin{tabular}[c]{@{}c@{}}4.20 \\ [-0.3em]($\uparrow$ 1.18)\end{tabular} 
       & \begin{tabular}[c]{@{}c@{}}97.20 \\ [-0.3em]($\uparrow$ 3.50)\end{tabular} 
       & \begin{tabular}[c]{@{}c@{}}0.84 \\ [-0.3em]($\downarrow$ 3.36)\end{tabular}\\
       [0.8em]
Socratic Models & \begin{tabular}[c]{@{}c@{}}6.92 \\ [-0.3em]($\uparrow$ 0.12)\end{tabular} 
       & \begin{tabular}[c]{@{}c@{}}4.92 \\ [-0.3em]($\uparrow$ 1.23)\end{tabular} 
       & \begin{tabular}[c]{@{}c@{}}97.22 \\ [-0.3em]($\uparrow$ 11.63)\end{tabular} 
       & \begin{tabular}[c]{@{}c@{}}14.29 \\ [-0.3em]($\uparrow$ 0.01)\end{tabular} \\
       [0.8em]
Retrieve-Socratic & \begin{tabular}[c]{@{}c@{}}7.00 \\ [-0.3em]($\uparrow$ 0.12)\end{tabular} 
       & \begin{tabular}[c]{@{}c@{}}5.89 \\ [-0.3em]($\uparrow$ 2.10)\end{tabular} 
       & \begin{tabular}[c]{@{}c@{}}96.88 \\ [-0.3em]($\uparrow$ 9.21)\end{tabular} 
       & \begin{tabular}[c]{@{}c@{}}17.65 \\ [-0.3em]($\uparrow$ 3.37)\end{tabular} \\
\bottomrule
\end{tabular}}
\vspace{-3mm}
\caption{Performance comparison on the ``shorter'' split. Arrows indicate relative differences compared to the ``short'' split.}
\vspace{-4mm}
\label{tab:shorter_split}
\end{table}

We observe a clear performance increase on the ``shorter'' split across summarization and ordering tasks, indicating that as temporal structure becomes more complex, models suffer substantial degradation even in relatively coarse-grained temporal reasoning. 
In addition, performance on temporal localization tasks remains consistently poor across all settings—Gemini-1.5-Flash’s result even declines on shorter inputs. This highlights that fine-grained temporal grounding is an intrinsic bottleneck, independent of sequence length or structure. The task requires pinpointing events at the granularity of a few seconds within videos spanning tens of minutes, a challenge exacerbated by token budget limits and information bottlenecks. General-purpose models are not well-optimized for this setting, and while recent methods show progress on shorter videos, performance on extended sequences remains severely constrained.

\noindent\textbf{Remarks.} 
(1) Model refusal rate: We observe a notable refusal rate from Gemini-1.5-Flash due to its built-in information security constraints, particularly when directly processing long-form video input. The refusal rates increase with input length, reaching 20.14\%, 27.08\%, and 30.56\% for short, medium, and long videos, respectively. Both the Socratic and Retrieve-Socratic methods, which convert video content into textual descriptions before prompting the model, achieve a 0\% refusal rate. LongVU does not exhibit any refusal behavior. (2) Temporal reasoning failure: In the multi-video summarization task, a significant portion of responses are invalid, even when timestamps are included in input context, and the query time is explicitly specified in the question (\eg, ``12:00 to 17:00''). Typical invalid responses include statements like ``There is no information available within the specified time period'', indicating a breakdown in the temporal understanding of the model. The failure rates of the multi-video summarization task are 55.28\%, 51.89\%, and 57.03\% for Gemini-1.5-Flash, Socratic Models and Retrieve-Socratic, respectively. 
LongVU demonstrates almost no temporal awareness, it consistently disregards the specified time window and instead summarizes the entire video content. Also, in the ordering task, its performance on short, medium, and long data drops drastically to 37.5\%, 1.7\%, and 0.69\%, respectively. Manual review of the outputs reveals that even on short data, LongVU frequently returns meaningless outputs like “0, 1, 2, 3” without establishing any meaningful temporal structure. Therefore, we only included LongVU in tasks that require global summaries or coarse-grained time reasoning (\ie, holistic summarization and ordering). Additional results and analysis are in the Appendix~\ref{appen:additional_results}.

\subsection{Summary of Key Findings}

\noindent\textbf{Temporal Reasoning: The Core Bottleneck.}
Evaluated models struggle with temporal reasoning, showing high failure rates in multi-video summarization and localization. This highlights the need for models with improved temporal alignment and contextual memory for long sequences.

\noindent\textbf{Textual Representations: A Path Towards Scalable Understanding.}
Our findings suggest that structured textual representations serve as a highly effective and scalable intermediate for ultra-long video understanding. As demonstrated by the Socratic and Retrieve-Socratic methods, converting raw long video inputs into textual forms significantly reduces model refusal rates and consistently enables more stable and improved performance. This strategy not only effectively alleviates the severe token limitations of current multi-modal models but also dramatically reduces computational overhead. This makes it a practical and highly scalable alternative for long-form video understanding, especially in scenarios where direct processing of extended video sequences remains challenging.

\noindent\textbf{Context Filtering: Balancing Efficiency and Completeness.}
The Retrieve-Socratic approach leverages rule-based temporal filtering to retain only task-relevant context, resulting in superior performance in temporal localization and summarization tasks. However, its relatively weaker performance in counting task suggests that aggressive pruning may omit essential contextual cues. Going forward, this insight underscores the need to design more adaptive retrieval-augmented generation (RAG) systems capable of balancing context filtering with comprehensive long-term information retention in ultra-long video understanding.

\section{Conclusion}
\label{sec:conclusion}

We introduce X-LeBench, the first benchmark for ultra-long egocentric video understanding, featuring a customizable life-logging simulation pipeline. By curating 432 video life logs from Ego4D, categorized into short, medium, and long durations. X-LeBench provides a diverse, structured dataset for long-form video analysis. It includes tasks in temporal localization, summarization, counting, and ordering, offering a rigorous evaluation framework. We hope to advance research in long-form egocentric video processing, fostering the development of more robust and temporally aware AI models.

\section{Limitations}
\label{sec:limitations}
Despite our efforts to construct an extremely long video dataset with contextually coherent activity contents, limitations remain due to the scarcity and insufficient diversity of available data. For details of videos, contextual inconsistencies exist within the dataset, such as different kitchen or office settings within the same data. However, since our focus is on long-term activity content, these discrepancies are beyond the scope of this work. In addition, while MBTI is employed in our simulation pipeline to introduce behavioral diversity, we acknowledge that it is a heuristic framework with limited scientific validity. Its inclusion serves a pragmatic purpose—to support structured variation in persona-driven daily plans—and is not intended to imply psychological rigor or generalizable personality modelling. 

Future work could involve expanding both the modalities and time spans of collected data through diverse devices and sources to enhance a comprehensive understanding of long-form human activity, supporting the development of more personalized tasks that emphasize long-term memory or preference-centric reasoning and predictive capabilities.

\bibliography{custom}

\appendix

\section{Persona Profile}
\label{appen:profile}
To simulate diverse egocentric behaviors, we generated 144 unique persona profiles by combining 16 MBTI personality types with 9 different location settings. Each profile serves as input for our dataset creation. Tab.~\ref{tab:persona_profile} presents a detailed example of a persona profile, illustrating its structure and content.

\begin{table*}[!t]
\centering
\renewcommand{\arraystretch}{1}
\resizebox{0.85\textwidth}{!}{%
\begin{tabular}{lp{16cm}}
\toprule
\textbf{Field} & \textbf{Description} \\ \hline
Location & UK \\ \hline
Personality Traits & 
\begin{itemize}
    \item MBTI Type: INFP
    \item Character Traits:
    \begin{itemize}
        \item Idealistic and creative, often daydreaming about a better world
        \item Highly empathetic and compassionate, with a strong sense of personal values
        \item Prefers deep, meaningful connections over a wide social circle
        \item Flexible and adaptable, but can be disorganized and easily overwhelmed by details
        \item Values authenticity and is highly attuned to the feelings of others
    \end{itemize}
\end{itemize} \\ \hline
Lifestyle & C wakes up around 7:30 AM and starts the day with meditation or journaling. Works from home as a freelance writer from 9:00 AM to 5:00 PM, with breaks for lunch and short walks. Evenings are spent reading, writing poetry, or engaging in creative projects, and bedtime is around 11:00 PM. \\ \hline
Hobbies & 
\begin{itemize}
    \item Writing poetry and short stories
    \item Reading literature and philosophy
    \item Practicing mindfulness and meditation
    \item Volunteering at local community centers
    \item Exploring nature and taking long walks in the countryside
\end{itemize} \\ \hline
Daily Agenda & 
\begin{itemize}
    \item Wake up at 7:30 AM
    \item Meditation or journaling from 8:00 AM to 8:30 AM
    \item Start work at 9:00 AM
    \item Lunch break from 1:00 PM to 2:00 PM
    \item End work at 5:00 PM
    \item Evening walk or light exercise from 6:00 PM to 7:00 PM
    \item Dinner and relaxation from 7:00 PM to 9:00 PM
    \item Reading or creative activities from 9:00 PM to 10:30 PM
    \item Go to bed at 11:00 PM
\end{itemize} \\ \hline

Daily Plan Chunks & \begin{itemize}
                              \item 07:30–08:00: Wake up and stretch
                              \item 08:05–08:35: Meditation or journaling
                              \item 08:40–09:10: Prepare a healthy breakfast and make tea
                              \item 09:15–09:45: Eat breakfast, clean the kitchen, and set up the workspace
                              \item 09:50–10:50: Start work as a freelance writer
                              \item 10:55–11:00: Continue working on writing projects
                              \item 11:05–11:20: Take a short break to stretch and hydrate
                              \item 11:25–13:00: Continue working on writing projects
                              \item 13:05–14:00: Lunch break and light reading
                              \item 14:05–16:00: Resume work and handle emails and client communications
                              \item 16:05–17:00: Continue handling emails and client communications
                              \item 17:05–17:35: End work and tidy up the workspace
                              \item 17:40–18:40: Evening walk or light exercise in the park
                              \item 18:45–20:00: Prepare and eat dinner, relax with some music
                              \item 20:05–21:35: Engage in creative activities (writing poetry, reading literature, or working on a personal project)
                              \item 21:40–22:30: Wind down with light reading or calming music
                            \end{itemize}                                                                                                                                         \\ \bottomrule
\end{tabular}%
}

\caption{Persona Profile: Tabular representation of an INFP individual’s daily life, lifestyle, hobbies, and activities.}
\label{tab:persona_profile}
\end{table*}

\section{Coarse-to-Fine Video Matching and Iterative Optimization}
\label{appen:satge3}
As described in Sec.~\ref{subsec:sim_pipeline}, the life-log simulation pipeline generates multiple daily activity chunks, which are then matched with videos from the video selection library to construct the dataset. Stage 3 follows a coarse-to-fine matching approach based on Equation~\ref{eq:simulation} with the implementation details outlined below:

\subsection{Coarse Matching (\textbf{$M_v(\cdot)$})}
For each activity chunk $C$, key information is extracted, including 
$C_t$ (time period), $C_{scene}$ (scene type), and $C_{scenario}$ (scenario). The $C_{scene}$ and $C_{scenario}$ attributes are inferred using GPT-4o based on the chunk’s textual description $C_{desc}$. Candidate videos are retrieved from the video selection library where the resource region matches the persona's predefined location. A video is included in the candidate set \textbf{$M_v(\cdot)$} when its time and scene type match the $C_t$ and $C_{scene}$, with an overlap between its scenarios and $C_{scenario}$ larger than 0.33 (\ie, the IoU of their scenarios). 
Note that, if no matching videos are found, the criteria are relaxed by lowering the overlap threshold to 0.2 or removing the regional constraint. This process produces the initial candidate set \textbf{$M_v(\cdot)$}.

\subsection{Fine Matching ($S(\cdot)$)}
From the candidate set \textbf{$M_v(\cdot)$}, a finer selection is made by computing the sentence similarity $S(\cdot)$ between the chunk’s textual description $C_{desc}$ and the textual summaries of the candidate videos. The two videos with the highest similarity scores are identified, and one is randomly selected to increase dataset diversity while maintaining relevance.

\subsection{Iterative Optimization}
Once a matching video is selected, its metadata is incorporated into the persona’s memory as a record of completed activities.
This updated memory, along with the persona’s predefined attributes, is provided to GPT-4o to perform a reflection process, enabling the model to refine and adjust the upcoming plan chunks to maintain logical coherence. 
The next plan chunk is then processed using the updated persona memory, and coarse-to-fine matching is repeated iteratively.
This process continues until all plan chunks are matched with corresponding videos, ensuring contextually coherent and realistic life-log simulations.

\section{Task Details}
\label{appen:tasks}
\begin{figure}[h]
  \centering
   \includegraphics[width=\linewidth]{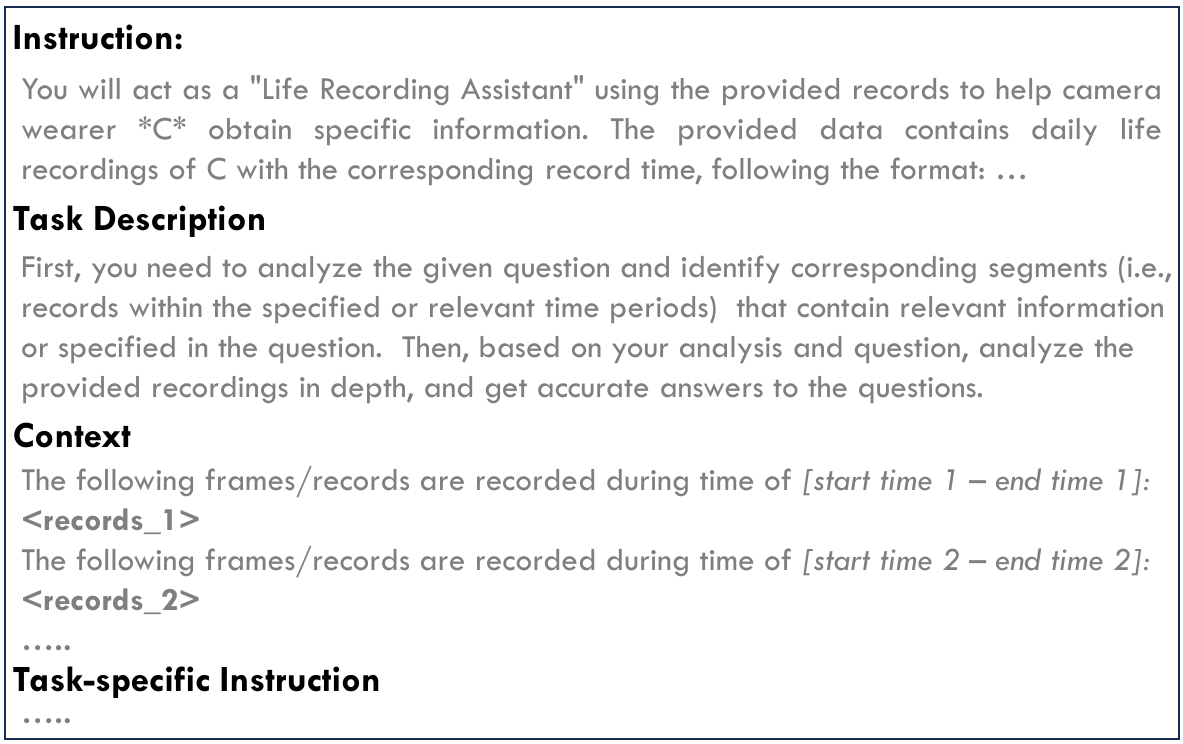}

   \caption{Example of input context.}
   \label{fig:context}
\end{figure}

As described in Sec.~\ref{subsection:tasks}, X-LeBench provides a comprehensive benchmark for daily activity-related tasks, comprising four major categories with eight subtasks. Here, we detail the task settings and the prompt templates used during evaluation. The prompt templates and example outputs are illustrated in Tab.~\ref{table:task_examples} and Fig.~\ref{fig:context}.

\subsection{Temporal Localization}

\noindent\textbf{Task Design and Annotation}: This task evaluates a model's ability to locate relevant time segments in ultra-long egocentric videos based on user queries. This task, akin to Episodic Memory~\cite{grauman2022ego4d} and Natural Language Video Localization~\citep{zhang2021natural, zhang2020span}, is uniquely challenging for ultra-long videos due to their extensive duration and diverse content. We divide this task into three subtasks based on the type of information being retrieved: objects, people, and moments. For object- and people-related retrieval, we adopt the annotations from the Episodic Memory task (specifically, the Natural Language Query subtask) in Ego4D. Specifically, we retrieve the annotations corresponding to the videos selected by the daily plan chunks (as generated in Sec.~\ref{subsec:sim_pipeline}), classify them into topics based on the query templates defined in Ego4D, and remapped the timestamps to align with the virtual time used in the daily plan. 

While Ego4D defines three query categories—objects, people, and place—we argue that the ``Place'' query template (``Where did I put X?'') fundamentally pertains to object location. Thus, we reclassify these queries under the ``object-related'' category. Additionally, queries without explicit template annotations were assigned to one of 13 predefined types using an edit-distance-based classification, followed by manual verification. For moment retrieval, annotations for this subtask were derived from the Moments Query subtask of the Episodic Memory task in Ego4D. 

\noindent\textbf{Evaluation Metrics}: Following previous works on temporal localization~\cite{grauman2022ego4d, zhang2021natural, zhang2020span}, we employed a top-$x$ recall with an intersection over union threshold (``recall@x, tIoU''). $x$ and $t$ are predefined variables that indicate the number of retrieved results we look at and the IoU threshold, respectively. Here, we set $x$ to 5 and $t$ to 0.3.

\subsection{Summarization}

\noindent\textbf{Task Design and Annotation}: Summarization is a fundamental task in numerous benchmarks~\cite{wang2024lvbench, zhou2024mlvu, ataallah2024infinibench}. In our context, it provides insights at varying granularities—specific chunks, predefined periods (\eg, morning, afternoon, evening), and holistic summaries. These diverse scopes challenge systems to summarize at multiple summary levels in ultra-long videos. Our summarization tasks involve generating hierarchical summaries for: Single-video (chunk-level) summaries; Multi-video summaries for predefined periods (morning, afternoon, evening\footnote{Morning: before 12 p.m., afternoon: 12 p.m. to 5 p.m., evening: after 5 p.m.}); And holistic (full-day) summaries. To generate summaries at these levels, we employed a hierarchical approach inspired by~\cite{islam2024video}. 
Single-video summaries are initially consolidated during the earlier stages of processing (as described in Sec.~\ref{subsec:sim_pipeline}). Then, multi-video summaries are produced by LLM-based aggregation, followed by holistic summaries generated in a similar manner.

\noindent\textbf{Evaluation Metrics}: We incorporate LLM-based evaluation metrics inspired by MLVU~\cite{zhou2024mlvu} to evaluate our summary tasks from the perspectives of 'Completeness' and 'Reliability'.

\subsection{Counting}

\noindent\textbf{Task Design and Annotation}: 
This task assesses a model's ability to identify, track, and count occurrences of specific actions across ultra-long videos. The goal is to test fine-grained aggregation capabilities over multiple time segments. Future extensions may include object/person counting.

To maintain annotation reliability, we limit the counting task to intervals with available annotations, as Ego4D annotation only covers a fraction of the total video duration. Unannotated actions could otherwise lead to incorrect counts. Specifically, we derive annotations from the Forecasting task in Ego4D, ensuring consistency. 

\noindent\textbf{Evaluation Metrics}: 
We employed a simple accuracy metric. Accuracy was calculated as the ratio of correctly counted actions to the total number of actions queried.

\begin{table*}[ht!]
\renewcommand{\arraystretch}{0.8}
\centering
\caption{Comparison between our method and random sampling. Left two columns are from our method; Right two columns are from random baseline.}
\scalebox{0.7}{
\begin{tabular}{ccccc}
\toprule
&  &  &  & \\[-10pt]
\textbf{Video Info.}    & \textbf{Video Summary} & \textbf{Timestamp} & \textbf{Video Summary} & \textbf{Video Info.}  \\
\hline
&  &  & & \\[-6pt]
Daytime \& Indoor & \begin{tabular}[c]{@{}c@{}}C washes hands and holds\\ the cupboard door in the bathroom.\end{tabular} & 06:35 & \begin{tabular}[c]{@{}c@{}}C talks, reads a book, arranges books,\\ draws, and walks inside a house.\end{tabular} & Daytime \& Indoor \\
\hline% \cline{2-6}
&  &   & \\[-6pt]
Daytime \& Indoor & \begin{tabular}[c]{@{}c@{}}C, X, Y, and O discuss\\ in a meeting room.\end{tabular} & 08:35 & \begin{tabular}[c]{@{}c@{}}C scrolls through the phone,\\ climbs a ladder, and fixes cables.\end{tabular} & Daytime \& Outdoor \\
\hline
&  &   & \\[-6pt]
Daytime \& Indoor & \begin{tabular}[c]{@{}c@{}}C dials a phone, drinks water,\\ and writes in a bedroom.\end{tabular} & 16:35 & C plays a drum set in a studio. & Nighttime \& Indoor \\
\hline
&  &   & \\[-6pt]
Nighttime \& Indoor & C reads a book on the bed in the bedroom. & 23:00 & C prepares avocado in the kitchen. & Daytime \& Outdoor \\
\bottomrule
\end{tabular}
}
\label{tab:trivial_baseline_comparison}
\end{table*}

\subsection{Ordering}
\noindent\textbf{Task Design and Annotation}: 
Ordering task evaluates a model’s capability to recognize temporal relationships between main activities across video records in ultra-long video life logs. Unlike prior studies focusing on fine-grained actions, we introduce the novel challenge of ordering events based on single-video summaries. Models are tasked with ordering shuffled single-video summaries of entire life logs, posing significant challenges to the system's temporal understanding ability in long-context inputs. 
This setup ensures the initial sequence of events is obscured, requiring the model to reconstruct the correct temporal order. 

We formulate the task as an open-ended sequence generation problem. Each video summary receives an index post-randomization, preventing numerical hints about the original order. The model is expected to output the correct order based on its understanding of the contents. 

\noindent\textbf{Evaluation Metrics}: We use ordering accuracy, defined as the ratio between the number of correctly predicted order and the total number of summaries.

\section{Dataset Quality Evaluation}
\label{appen:quality}
X-LeBench will be made publicly available, users can access and download videos for research under the Ego4D license (see \href{https://ego4d-data.org/docs/}{Ego4D website}).
To validate the quality of our dataset, we conducted a human evaluation focusing on \textbf{realism} and \textbf{contextual consistency}. These key aspects are assessed to ensure the dataset accurately reflects realistic human behavior patterns and maintains logical continuity across activities throughout the day. Evaluators concentrated on the content of the activities, disregarding minor inconsistencies (\eg, clothing or environmental details) that do not affect the overall activity realism. The implementation details of this evaluation are outlined in the following sections.
\subsection{Evaluation Criteria}
The evaluation criteria are defined as follows:
\begin{itemize}
    \item \textbf{Realism.} Each record is assessed for temporal and content alignment with real-world daily activities. Mismatches, such as daytime footage with nighttime timestamps, are considered unrealistic. Evaluators should score all records from 1 (very unrealistic) to 5 (highly realistic).
    \item \textbf{Contextual Consistency.} This criterion evaluates the logical flow and smoothness between consecutive records. Abrupt transitions, such as jumping from office work to outdoor activities, are considered a lack of consistency. Evaluators should score all records from 1 (very poor consistency) to 5 (strong sequence). This ensures the dataset reflects realistic daily human behavior.
\end{itemize}

\begin{figure*}[h]
  \centering
   \includegraphics[width=0.8\textwidth]{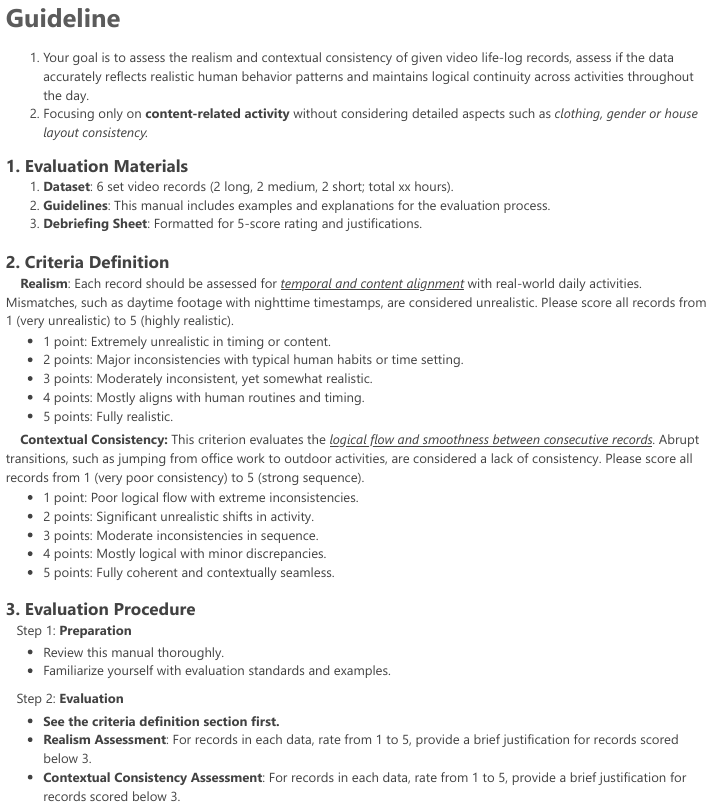}
   \caption{The content of guideline.}
   \label{fig:guideline}
\end{figure*}

\subsection{Evaluation Setting}
Given the extremely long duration of the dataset, which makes full evaluation cost-prohibitive, we select 6 data samples (two each from the long, medium, and short categories), consisting of 56 videos, totalling 31.9 hours. The evaluators are provided with a detailed evaluation manual and the original video data. Each evaluator independently assesses the video records and records their results on a formatted debriefing sheet. Kendall's W~\cite{abdi2007kendall} is computed to assess the inter-rater reliability (IRR) of the evaluations.

\subsection{Evaluation Procedure}
\noindent\textbf{Evaluator Training.} Evaluators are trained using detailed guideline (Fig.~\ref{fig:guideline}) that include evaluation standards, examples, and formatting instructions. 

\noindent\textbf{Evaluation.} Three independent evaluators assess all provided data following the guidance. Evaluators score all records from 1 to 5 for realism and contextual consistency, with a brief justification for records scored below 3.

\noindent\textbf{Consensus Evaluation.} Compute inter-rater reliability (IRR) using Kendall's W for the scores of 56 records to ensure reliability between evaluators. Resolve discrepancies through discussion and re-evaluation. Here, the evaluations achieve 0.536, which suggests a moderate level of agreement among the raters.

\noindent\textbf{Reporting.} The final report presents the mean score for all evaluators.

\section{Additional Results}
\label{appen:additional_results}
\subsection{Comparison with Trivial Baseline}
Our proposed pipeline is designed to maintain temporal coherence and contextual consistency by integrating multi-dimensional information (time, scene, and content) via life-logging simulation pipeline. To validate the effectiveness of our method, we compare it with a trivial baseline that randomly samples videos. Table~\ref{tab:trivial_baseline_comparison} shows the comparison between the two methods.

As can be observed, our approach retrieves videos with contextually appropriate timestamps and consistent content, closely mirroring real-life life-logging scenarios. In contrast, random sampling results in time-inconsistent and semantically disjointed videos, further validating the effectiveness of our method. 

\subsection{Evaluation on LongVU}
To extend our evaluation, we tested LongVU~\cite{shen2024longvu}, a state-of-the-art open-source model tailored for long video understanding. Results are summarized in Table~\ref{tab:longvu_results}.

\begin{table}[h!]
\renewcommand{\arraystretch}{0.8}
\centering
\caption{Performance of LongVU.}
\scalebox{0.65}{
\begin{tabular}{ccc}
\toprule
% &  &  &  & &  \\[-10pt]
\textbf{Data Category} & \textbf{Holistic Summary (10)} & \textbf{Ordering (\%)} \\
\midrule
% &  &  &   \\[-6pt]
Short & 5.21 & 37.5 \\
\hline% \cline{2-6}
&  &    \\[-6pt]
Medium & 4.81 & 1.70 \\
\hline% \cline{2-6}
&    &   \\[-6pt]
Long & 4.73 & 0.69 \\              
\bottomrule
\end{tabular}
}
\label{tab:longvu_results}
\end{table}
Our manual review of outputs shows LongVU struggles with temporal reasoning. Even with explicit cues (\eg, ``frames 0–500 are recorded from 13:00–13:10''), it still outputs summaries covering the full video. In ordering tasks, LongVU frequently outputs invalid sequences such as ``0, 1, 2, 3'' or ``100, 100, 100, 100'', especially for long videos. Consequently, we only evaluated LongVU on tasks that require global information or coarse-grained temporal understanding (\ie, holistic summarization and ordering). Its performance significantly drops as video length increases, underscoring the limitations of current open-source models in ultra-long video understanding and temporal reasoning.

\end{document}